# CapsGAN: Using Dynamic Routing for Generative Adversarial Networks

**Raeid Saqur**
Department of Computer Science
University of Toronto
`raeidsaqur@cs.toronto.edu`

**Sal Vivona**
Department of Computer Science
University of Toronto
`vivona@cs.toronto.edu`

## Abstract

In this paper, we propose a novel technique for generating images in the 3D domain from images with high degree of geometrical transformations. By coalescing two popular concurrent methods that have seen rapid ascension to the machine learning zeitgeist in recent years: GANs (Goodfellow et. al.) and Capsule networks (Sabour, Hinton et. al.) - we present: **CapsGAN**. We show that CapsGAN performs better than or equal to traditional CNN based GANs in generating images with high geometric transformations using rotated MNIST. In the process, we also show the efficacy of using capsules architecture in the GANs domain. Furthermore, we tackle the Gordian Knot in training GANs - the performance control and training stability by experimenting with using Wasserstein distance (gradient clipping, penalty) and Spectral Normalization. The experimental findings of this paper should propel the application of capsules and GANs in the still exciting and nascent domain of 3D image generation, and plausibly video (frame) generation.

## 1   Introduction

Deep learning has made significant contributions to areas including natural language processing and computer vision. Most accomplishments involving deep learning use supervised discriminative modeling. However, the intractability of modeling probability distributions of data makes deep generative models difficult. Generative adversarial networks (GANs) [4] help alleviate this issue through setting a Nash Equilibrium between a generative neural network model (Generator) and a discriminative neural network (Discriminator). The discriminator is trained to determine whether its input is from a real data distribution or a fake distribution that was generated by the generative network.

Since the advent of GANs, many applications and variants have risen. Most of its applications are inspired by computer vision problems, and involve image generation as well as (source) image to (target) image style transfer. Although GANs have been proven to be successful in the space of computer vision, most use Convolutional Neural Networks (CNN) as their discriminative agents. Through convolutions and pooling, CNN are able to abstract high level features (e.g. face, car) from lower level features (e.g. nose, tires). However, CNNs lack the ability to interpret complicated translational or rotational relationships between lower features that make up higher level features [14] - they lack '*translational equivariance*'. CNNs attempt to solve this relational issue through imposing translational invariance while increasing its field of view through max or average pooling. However, these pooling operations are notorious for losing information that can be potentially essential for the discriminator's performance.

**Capsules** were introduced in [7]. Capsules offers a better model for learning hierarchical relationships of lower features when inferring higher features [14]. These capsules are able to decode aspects such as pose (position, size, orientation), deformation, velocity, albedo, hue, texture etc. Based on the limitations of CNN's, we propose using Capsule Networks as the discriminator network for a simple GAN framework. This will allow the discriminator to abide by inverse-graphics encoding. As a result, this will encourage the generator to generate samples that have more realistic lower to higher feature hierarchies.

## 2   Related Work

Training GANs architecture has been notoriously difficult [15] - as they suffered from instability due to 'mode collapse'. The original GANs implementation using feed-forward MLP did not perform well generating images in complex datasets like the CIFAR-10, where generated images suffered from being noisy and incomprehensible. An extension to this design using Laplacian pyramid of adversarial networks [3] resulted in higher quality images, but they still suffered from wobbly looking objects due to noise emanating from chaining multiple models. In order to circumvent the issues and make training feasible, a plethora of techniques were proposed in the last few years, most of which extended or altered existing popular architectural designs. Notable and popular ones include 'DCGAN' [13] and 'GRAN' [8] that are based off of CNN and RNN classes of architecture respectively. Following the trend, [12] proposed 'Conditional GANs', which includes an architectural change to GANs where the discriminator also acts as a classifier for class-conditional image generation.

In this paper, we used Radford et. al.'s [13] work in the sense that we propose to replace the CNN based architecture of '**DCGAN**' with capsule networks. We perform experiments and validation tests using the MNIST dataset, as it is the dominant test dataset for majority of the papers in this domain, allowing us to juxtapose our results with preceding architectures. As future work, the performance measure can be tested on CIFAR-10,100 datasets (*discussed more in detail in 6*).

## 3   CapsGAN

This section details the architecture and algorithm of the baseline DCGAN and proposed CapsGAN used for the rotated MNIST dataset.

### 3.1   Architecture

For the **baseline**, the DCGAN architecture was used. The following Figure 1 shows the topological architecture of the baseline GAN architecture used.

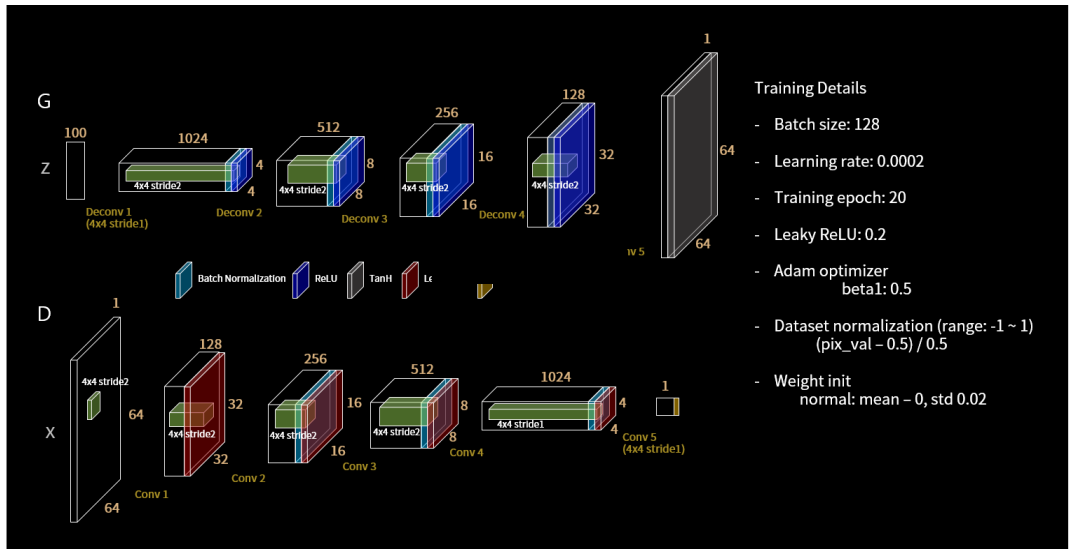

Figure 1: Topological baseline Architecture using DCGAN

For **CapsGAN**, the model follows a vanilla GAN framework, dually training both a generator and discriminator network. The discriminator is represented as a capsule network which uses dynamic routing, with its implementation being a minor variant of the one featured in the original paper [14]. The key difference is the final layer which is one 16 dimensional capsule, while the original architecture featured $10 \times 16$ dimensional capsules. This was done, as the discriminator's output should represent the probability of whether the input image was sampled from the real or generated distribution. Thus, this binary information requires one 16-D vector capsule, with its length being correlated to the probability of the input image being sampled from the real data distribution. The dynamic routing process occurs between the primary capsule layer and the output digit capsule. A high-level illustration is shown below in Figure 2

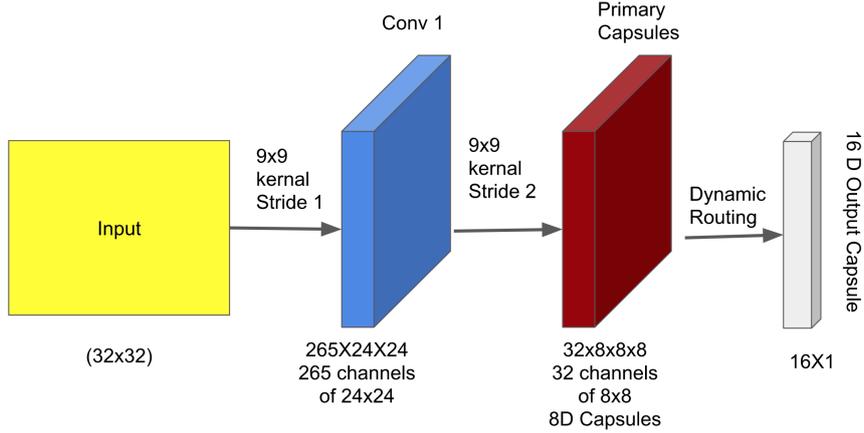

Figure 2: Capsule Discriminator with one output capsule

The generator involves a 2D transpose convolutional network, which is illustrated in the figure 1. The generator network was inspired from the DCGAN framework [13]. It incorporates 5 convolution transpose layers, which are responsible for up-sampling a 128 dimensional vector with each element being independently sampled from a normal distribution of mean 0 and standard deviation 1. Batch Normalization is also applied between each of the deconvolution layers. This is done to regularize the generative network, which would in turn increase training stability [9]. Rectified linear units are used as the nonlinearity between all layers with the exception of the final layer. The nonlinearity is applied to the output prior to batch normalization. A 'tanh' nonlinearity function is then applied to the final output of the up-sampled image.

### 3.2 Loss Function

While constructing the network's framework, **three types** of loss functions were considered. The first two were binary cross entropy, and mean squared error (MSE). However, during experiments, using these loss functions did not allow the CapsGAN to converge. For the DCGAN baseline, binary cross entropy was used for training. The CapsGAN loss function was chosen to be the margin loss as introduced in the original dynamic routing paper. Using this loss encouraged the model to converge without any mode collapse.

$$v_i = CapsD(x_i) \tag{1}$$

$$Loss = T_i \, max(0, m^+ - \|v_i\|^2) + \lambda(1 - T_i) \, max(0, \|v_i\| - m^-)^2 + \beta R_{loss} \tag{2}$$

Where $T_i$ is the target value (0 is fake, 1 is real), $v_i$ is the final output capsule vector, and $x_i$ is the input. Through hyper-parameter search, $\lambda$, $\beta$, $m^-$, $m^+$ were set to 0.5, 0.005, 0.1 and 0.9 respectively. The discriminator and generator optimized their weights based on the total loss function over the entire batch of images.

$$D^* = \max_w \{ (Loss(CapsD(x_i), T_i^{real} = 1) + \max_w (Loss(CapsD(x_i), T_i^{fake} = 0) \} \quad (3)$$

$$G^* = \max_w \{ Loss(CapsD(G(z_i)), T_i^{real} = 1)\} \quad (4)$$

In addition to margin loss, reconstruction loss ($R_{loss}$) was also added. This was inspired by the original dynamic routing paper which uses fully connected layers to reconstruct image inputs from the output capsule. This was included in the loss function in order to increase the quality of the capsule network classifier.

### 3.3 Algorithm

The following algorithm 1 shows the high-level training algorithm for CapsGAN[1].

---
**Algorithm 1** CapsGAN algorithm
---
1: **arguments**: Generator $G_\theta$, Discriminator $CapsD_\phi$
2: **initialize** The networks and other pertinent hyper-parameters ▷ number of epochs, LR etc.
3: **procedure** PRE-TRAIN DISCRIMINATOR
4:     **for** D-iters steps **do** ▷ Number of pre-training steps, default = 1
5:        Sample minibatch of *m* noise samples $\mathbf{Z} = \{z^{(1)}, ..., z^{(m)}\}$
6:        Sample minibatch of *m* samples $\mathbf{X} = \{x^{(1)}, ..., x^{(m)}\}$ from $p_{data}(x)$
7:        Update the discriminator, CapsD, using margin loss.
8: **procedure** ADVERSARIAL TRAINING($G, CapsD$)
9:     **for** number of epochs **do**
10:        **for** number of training steps **do**
11:           Sample minibatch of *m* noise samples $\mathbf{Z} = \{z^{(1)}, ..., z^{(m)}\}$
12:           Sample minibatch of *m* samples $\mathbf{X} = \{x^{(1)}, ..., x^{(m)}\}$ from $p_{data}(x)$
13:           Update discriminator by minimizing total loss: ▷ Train the discriminator
14:              $fake\_loss \leftarrow CapsD(G(Z))$
15:              $real\_loss \leftarrow CapsD(X)$
16:              $total\_loss \leftarrow real\_loss - fake\_loss$
17:           Update generator by minimizing fake loss: ▷ Train the generator
18:              $fake\_loss \leftarrow CapsD(G(Z))$
---

## 4 Experiments

The model was trained on the MNIST[2] dataset due to its simplicity. This involved tracking the proposed loss function over the number of total trained batches, where each batch contained 32 images with a width and height of 32 pixels. Prior to experimentations, grid hyper parameter search was performed for both the Capsule network as well as the baseline DCGAN model. For more valid comparisons, the number of parameters within the experimental and baseline model was controlled.

### 4.1 Methods

Both the CapsGAN as well as the baseline were trained until convergence of the loss functions. This typically involved training over 1000 to 2000 batches until model convergence.

In order to improve training stability, several additions were included. Specifically, these additions included spectral normalization [11], and incorporating Earth-Mover (EM) or Wasserstein distance for loss function optimizations, as was proposed and outlined by Arjovsky et. al.[1]. Experiments were done with both weight-clipping of gradients like the original paper, and also adding a gradient-penalty term as a suggested improvement by Gulrajani [5].

---
[1] https://github.com/raeidsaqur/CapsGAN
[2] http://yann.lecun.com/exdb/mnist/

Using grid hyper-parameter search, a single model was selected to represent the baseline and capsule GAN experimental model. After hyper-parameter search, the models were trained until 2000 batch iterations. However, both models appeared to converge at around 1000 iterations. Thus the results in the experiments exhibit the first 1000 iterations of both the baseline and the CapsGAN. Because it is not possible to apply inception scores to the MNIST dataset, the models were compared qualitatively based on their generated images as shown in Table 1 and Table 2. Table 1 represents both models being trained on images that have no prior geometric transformation. The models were also trained in two additional scenarios where geometric rotations were applied. This involved rotating the images based on its center axis after sampling an angle from a uniform distribution. The angular bounds of the uniform distribution represented the different training scenarios. The two geometrical rotations were uniformly sampled to have a maximum angle of 15 and 45 degrees. The performance of both the models are then qualitatively compared based on their generative images.

## 5 Results and Evaluation

### 5.1 MNIST

The experiments include observations for three different training scenarios. The first scenario is not rotated, while the other two scenarios are randomly rotated up to 15 and 45 degrees from its true orientation. This was done to help measure both of the model's sensitivity to geometric transformations. In the first experiment involves observing the loss function over the number of iteration batches for the CapsGAN. This plot will show the convergence properties of the loss function proposed in 3.2. The second part of the experiment involves qualitative comparisons between the reconstructions of the CapsGAN and DCGAN baseline. These reconstructions are visualized over numerous stages of training to show the qualitative convergence of the reconstructed images.

#### 5.1.1 Loss Convergence of CapsGAN

The following two figures 3 and 4 show the loss of discriminator and generator respectively during the training phase:

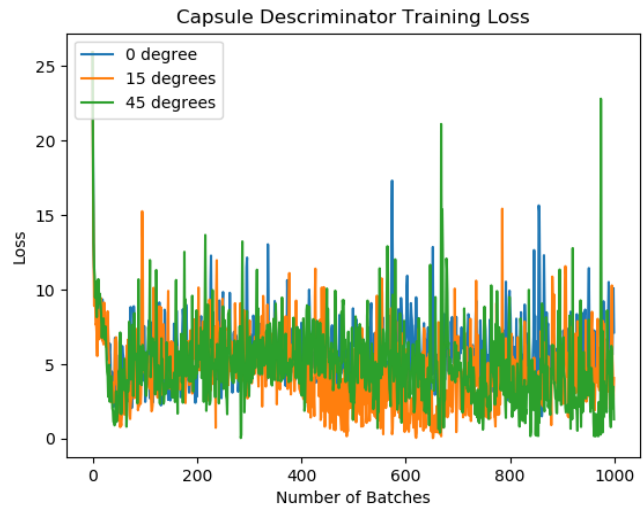

Figure 3: Loss of capsule based discriminator over the number of trained batches.

As shown in Figure 3, the Capsule discriminator rapidly converges at around 100 iterations, and continues to fluctuate heavily. Interestingly, the geometric rotations do not appear to affect the capsule networks' convergence. However, the scenario with up to 45 degrees of rotation involves the largest local variations and fluctuation peaks. Nonetheless, despite these local variations, the global trend seems to continue to stabilize while avoiding divergence. For the generator's loss function, a similar trend is exhibited as shown in Figure 4. The overarching trend is downward and decreasing, indicating improvement in the generator's loss function during training. However, the convergence also exhibits highly variant local fluctuations. In future work, there should be more focus on reducing the variance observed in this study. To help stabilize the convergence, '*Spectral Normalization*' [11] and *Wasserstein distance* [1] was implemented separately as well as in combination. Unfortunately, these implementations **did not** appear to improve the results. This is perhaps due to a lack of hyper-parameter tuning with the inclusion of these implementations. Thus for future work, extensive investment must be done for hyper-parameter tuning. This aspect of the experiment was partially hindered from the lack of available local and remote computational resources.

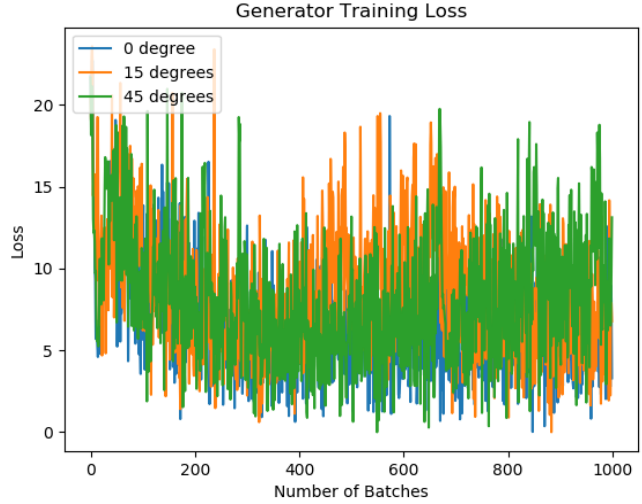

Figure 4: Loss of capsule based generator over the number of trained batches.

| Batch Iterations. | 500 | 1000 | 2000 |
|---|---|---|---|
| CAPS-DCGAN (0 degrees) | | | |
| DCGAN (0 degrees) | | | |

Table 1: CAPS-DCGAN and Baseline DCGAN Training on Non-Rotated Images.

### 5.1.2 Reconstruction of Non-Transformed MNIST

The first training scenario involves MNIST digits that are not rotated or transformed. Table 1 shows the image reconstructions from the baseline as well as the CapsGAN models. Through a qualitative comparison, it appears the CapsGAN generator converges to higher quality images than the DCGAN. This is shown at batch 2000 where the CapsGAN generated digits appear to be finer and more defined in comparison to the baseline. In order to further prove this claim, more quantitative measures should be done in future experiments. However qualitatively, it appears that CapsGAN converges to higher quality images while training on the same number of batches compare to the DCGAN baseline.

| Batch Iterations. | 300 | 500 | 1000 |
|---|---|---|---|
| CAPS-DCGAN (15 degrees) | 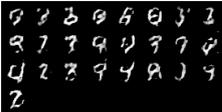 | 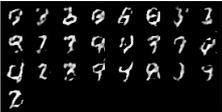 | 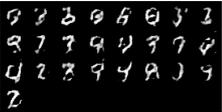 |
| DCGAN (15 degrees) | 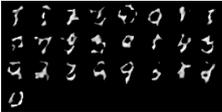 | 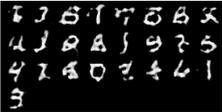 | 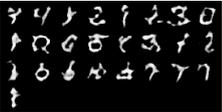 |
| CAPS-DCGAN (45 degrees) | 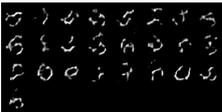 | 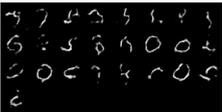 | 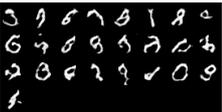 |
| DCGAN (45 degrees) | 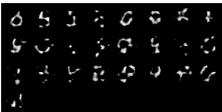 | 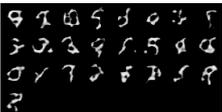 | 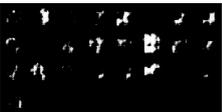 |

Table 2: CAPS-DCGAN and Baseline DCGAN training on rotated images that are rotated at angle that is sampled from a uniform distribution between the positive and negative value of the angle presented. Results show that for small angles, both models preform similar in their reconstructions. Larger angles result in divergence in the baseline at 1000 iterations.

### 5.1.3 CAPS-DCGAN and Baseline DCGAN Training on Rotated images.

As shown in Table 2, both the baseline and CapsGAN were trained on geometrically rotated data, as described in section 4. The table shows generated images from the baseline and CAPSGAN architecture for images rotated up to 15 degrees as well as for 45 degrees.

For 15 degree rotations, both models appear to show similar results at 300 and 500 iterations. However, at 1000 iterations, the CAPS-DCGAN exhibits higher quality generated digits compared to the baseline. This is indicated by more inappropriate stroke artifacts in the baseline images than the CapsGAN.

CapsGAN's advantage is more explicit with higher degree of geometric transformations applied to the image - as is illustrated by the experiment results with the images rotated 45 degrees. This is specifically shown at 1000 iterations where the baseline's reconstructions are shown to be unstable, while remaining stable for the CapsGAN. These results were aligned with expectations as Capsule networks are intended to be more robust to geometric transformations. This geometric adversity is due to the capsule's inverse graphics properties as discussed in the original paper [14].

| Categories | Batch Samples |
|---|---|
| Animals (id=0) | 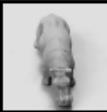 |
| Humans (id=1) | 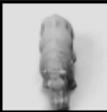 |
| Airplanes (id=2) | 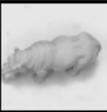 |
| Trucks (id=3) | 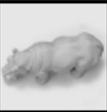 |
| Cars (id=4) | 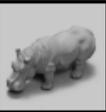 |

Table 3: Small-NORB dataset real dataset samples down-sampled to $64 \times 64$.

## 5.2 SmallNORB

The smallNORB [10] is a staple dataset for testing efficacy of generative models in the 3D domain. It is organized in gray-level stereo images of 5 classes of toys (please see Table 3). The multivariate geometric transformations applied to each of the images make it a quintessential dataset to benchmark. Each category is pictured at 9 elevations, 6 lighting conditions (notice the shadows across the samples), and 18 different azimuths. Training and test sets contain 24,300 stereo pairs of $96 \times 96$ images.

### 5.2.1 Baseline DCGAN Training on SmallNORB

The original images were downsampled to $64 \times 64$ pixels. The baseline architecture used was similar to the one in Figure 1. Batch size of 8 and 64 were used with batch normalizations. For the generator, random normal noise vector of size 100 was used as generator's input data. No pre-training was performed for the generator, and was left for future experiments. Lipschitz continuity was enforced to improve training stability by using gradient clipping after each gradient update[1].

### 5.2.2 Reconstruction using Baseline DCGAN

The following Figure 4 exemplifies the results obtained using our baseline architecture. Here we showcase only the '*cars*' category for brevity, as similar trends in results were observed for all the other categories. Batches of size 8 from the real data and generated data are juxtaposed at random training iteration intervals.

As can be seen, the generic trend shows convergence and the generated images gradually look sharper. However, due to the high degree of geometric transformations (lighting, elevation, azimuth, rotation), the baseline generator struggles to pin-point, i.e. converge on specific features.

This baseline experiment elucidates the weakness of traditional convolutional architecture based GANs to capture geometric transformations in images, and consequently in video frame generations.

| Category: 'Cars' {Data}_{iters} | **Samples** |
|---|---|
| Real_3000 | 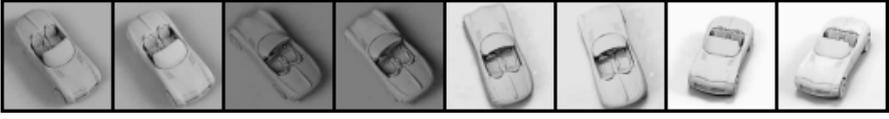 |
| Generated_3000 | 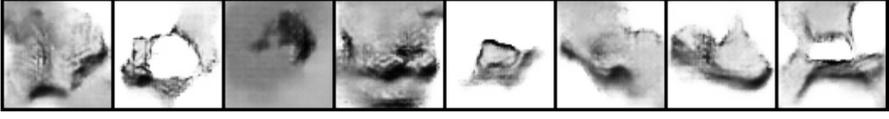 |
| Real_4700 | 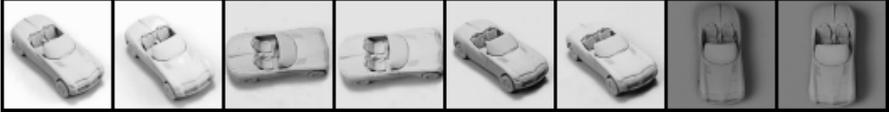 |
| Generated_4700 | 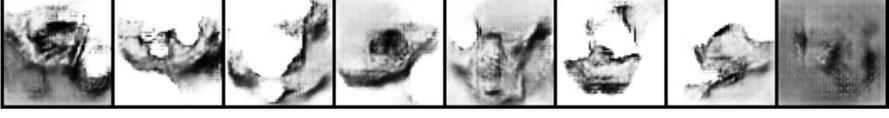 |
| Real_7000 | 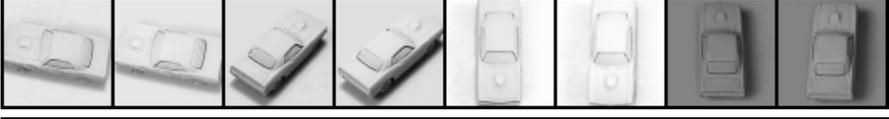 |
| Generated_7000 | 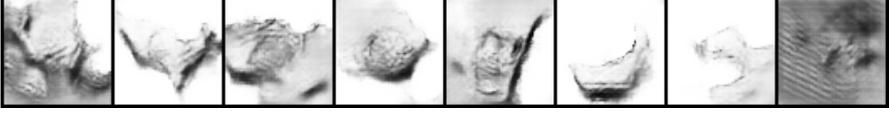 |
| Real_8200 | 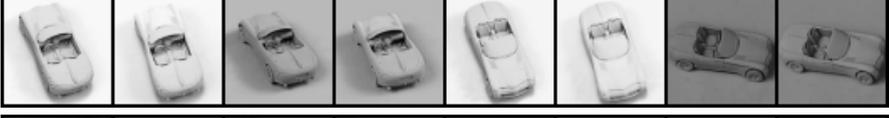 |
| Generated_8200 | 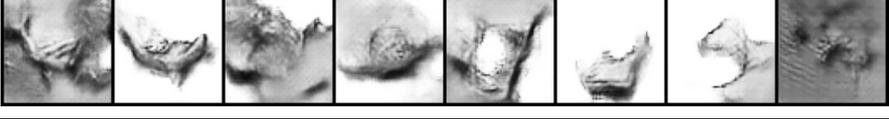 |
| Real_11250 | 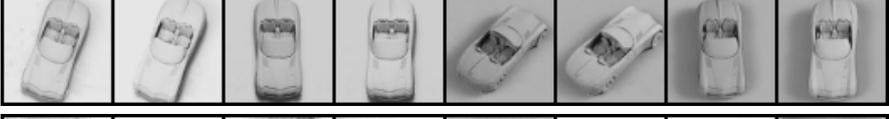 |
| Generated_11250 | 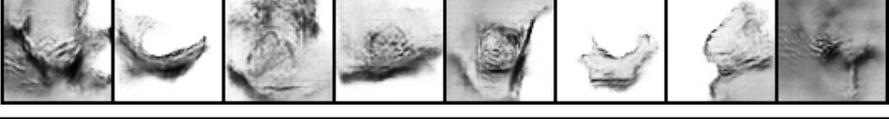 |

Table 4: Reconstruction of 'cars' category toy images using baseline DCGAN architecture. The left-most column indicates the type of data (real or generated) and the following number indicates the training iteration.

### 5.2.3 Proposed Architecture using Capsules

Our proposed architectural change for the discriminator is shown in the Figure 5 below. It is closely aligned with the 'capsule routing matrix' structure as outlined in [6].

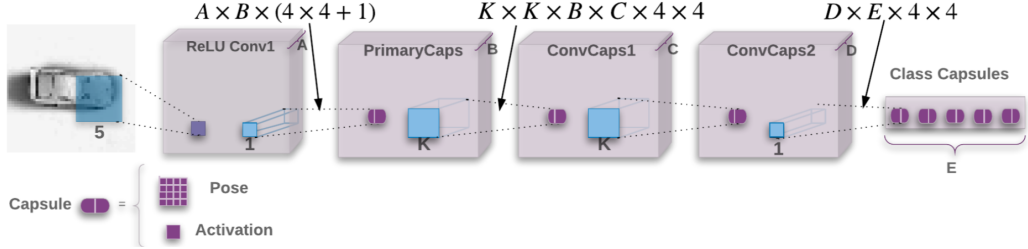

Figure 5: Proposed CapsGAN discriminator architecture for smallNORB dataset.

The model has a initial $5 \times 5$ convolutional layer with 32 channels (A), with a stride of 2 and ReLU non-linearity. All other layers are capsules. There are 32 primary capsules (B) and $4 \times 4$ pose of each of these capsules is a learned linear transformation from the lower layers. The primary capsules are followed by two $3 \times 3$ convolutional capsule layers (K), each with 32 capsule types (C=D=32) with strides of 2 and one, respectively. The last layer of convolutional capsules is connected to the final capsule layer which has one capsule per output class.

### 5.2.4 Results using CapsGAN

Table 5 shows reconstructed samples using the CapsGAN architecture.

| Category | Reconstructed Samples |
| --- | --- |
| Cars |  |
| Planes |  |

Table 5: Reconstruction of '*cars*' and '*planes*' categorical toy images using CapsGAN architecture.

## 6 Future Work and Limitations

Experiments and results from this paper show CapsGANs to be more robust to geometric transformations than traditional deep CNN based approaches. However, the simplistic image dataset (MNIST) used is a limitation, and more experiments using complex datasets are needed. The main comparative advantage of CapsGAN lies in its superior ability in capture geometric transformations, and thus results using smallNORB (or similar datasets) would certainly reinforce the notion.

Training stability during loss function optimization requires more work as loss convergence exhibited high variance. There is heavy research on GAN training stability, and newer, improved techniques can be explored to alleviate this issue. This may include doing more intensive hyper-parameter searches in order to find a more stable model. This can be encouraged through moving away from grid search methods and more towards Bayesian hyper-parameter search methods [2].

In this paper, we have established groundwork by illustrating the weakness of CNN architecture based GANs in generating images in the 3D domain. As an immediate future work, we will try and replicate the experiment using our proposed capsule based architecture. A more optimistic goal is to apply capsule networks to 3D generative modeling that is similar to the framework of the 3D-GAN [16]

Additionally, it is expected that future work would include more quantitative measures for comparing between the CapsGAN and a diverse set of baselines. It is understandable that although qualitative analysis of the generative 3D construction is intuitive, it does not capture the degree of the difference between the two model's performances.

## 7 Conclusion

In this paper, we have shown the efficacy and validity of using capsule networks in designing GAN architectures. We have also shown the promise of such GANs' superior capability in generating images from geometrically transformed dataset. Finally, we explored the Gordian Knot in GANs: the training stability - by performing ablation studies and experimentation with some of the more popular techniques proposed in recent machine learning academic literature.

We hope that our work in this paper will ignite interest and further work in the related domains of 3D image generation and the exciting and relatively unexplored domain of video generation.

## References

[1] Martin Arjovsky, Soumith Chintala, and Léon Bottou. Wasserstein gan. *arXiv preprint arXiv:1701.07875*, 2017.

[2] James S Bergstra, Rémi Bardenet, Yoshua Bengio, and Balázs Kégl. Algorithms for hyper-parameter optimization. In *Advances in neural information processing systems*, pages 2546–2554, 2011.

[3] Emily L Denton, Soumith Chintala, Rob Fergus, et al. Deep generative image models using a laplacian pyramid of adversarial networks. In *Advances in neural information processing systems*, pages 1486–1494, 2015.

[4] Ian Goodfellow, Jean Pouget-Abadie, Mehdi Mirza, Bing Xu, David Warde-Farley, Sherjil Ozair, Aaron Courville, and Yoshua Bengio. Generative adversarial nets. In *Advances in neural information processing systems*, pages 2672–2680, 2014.

[5] Ishaan Gulrajani, Faruk Ahmed, Martin Arjovsky, Vincent Dumoulin, and Aaron C Courville. Improved training of wasserstein gans. In *Advances in Neural Information Processing Systems*, pages 5769–5779, 2017.

[6] Geoffrey Hinton, Nicholas Frosst, and Sara Sabour. Matrix capsules with em routing. 2018.

[7] Geoffrey E Hinton, Alex Krizhevsky, and Sida D Wang. Transforming auto-encoders. In *International Conference on Artificial Neural Networks*, pages 44–51. Springer, 2011.

[8] Daniel Jiwoong Im, Chris Dongjoo Kim, Hui Jiang, and Roland Memisevic. Generating images with recurrent adversarial networks. *arXiv preprint arXiv:1602.05110*, 2016.

[9] Sergey Ioffe and Christian Szegedy. Batch normalization: Accelerating deep network training by reducing internal covariate shift. *arXiv preprint arXiv:1502.03167*, 2015.

[10] LeCun Jie Huang. The small norb dataset, v1.0.

[11] Takeru Miyato, Toshiki Kataoka, Masanori Koyama, and Yuichi Yoshida. Spectral normalization for generative adversarial networks. *arXiv preprint arXiv:1802.05957*, 2018.

[12] Augustus Odena, Christopher Olah, and Jonathon Shlens. Conditional image synthesis with auxiliary classifier gans. *arXiv preprint arXiv:1610.09585*, 2016.

[13] Alec Radford, Luke Metz, and Soumith Chintala. Unsupervised representation learning with deep convolutional generative adversarial networks. *arXiv preprint arXiv:1511.06434*, 2015.

[14] Sara Sabour, Nicholas Frosst, and Geoffrey E Hinton. Dynamic routing between capsules. In *Advances in Neural Information Processing Systems*, pages 3859–3869, 2017.

[15] Tim Salimans, Ian Goodfellow, Wojciech Zaremba, Vicki Cheung, Alec Radford, and Xi Chen. Improved techniques for training gans. In *Advances in Neural Information Processing Systems*, pages 2234–2242, 2016.

[16] Jiajun Wu, Chengkai Zhang, Tianfan Xue, Bill Freeman, and Josh Tenenbaum. Learning a probabilistic latent space of object shapes via 3d generative-adversarial modeling. In *Advances in Neural Information Processing Systems*, pages 82–90, 2016.

# Appendices

## A  Capsule Algorithms & Formulae

### A.1  Squashing Function

$$v_j = \frac{||s_j||^2}{(1+||s_j||^2)} \frac{s_j}{||s_j||} \quad (5)$$

Here, $v_j$ is the vector output of capsule j and $s_j$ is its total output.

### A.2  Margin Loss

The following equation shows the margin-loss used for digit existence:

$$L_k = T_k \, max(0, m^+ - ||v_k||)^2 + \lambda(1-T_k) \, max(0, ||v_k|| - m^-)^2 \quad (6)$$

• Softmax Function

$$c_{ij} = \frac{exp(b_{ij})}{\sum_k exp(b_{ik})} \quad (7)$$

### A.3  Routing Algorithm

---
**Procedure 1** Routing algorithm.
---
1: **procedure** ROUTING($\hat{u}_{j|i}, r, l$)
2:     for all capsule $i$ in layer $l$ and capsule $j$ in layer $(l+1)$: $b_{ij} \leftarrow 0$.
3:     **for** $r$ iterations **do**
4:         for all capsule $i$ in layer $l$: $\mathbf{c}_i \leftarrow \texttt{softmax}(\mathbf{b}_i)$          ▷ softmax computes Eq. 3
5:         for all capsule $j$ in layer $(l+1)$: $\mathbf{s}_j \leftarrow \sum_i c_{ij}\hat{\mathbf{u}}_{j|i}$
6:         for all capsule $j$ in layer $(l+1)$: $\mathbf{v}_j \leftarrow \texttt{squash}(\mathbf{s}_j)$          ▷ squash computes Eq. 1
7:         for all capsule $i$ in layer $l$ and capsule $j$ in layer $(l+1)$: $b_{ij} \leftarrow b_{ij} + \hat{\mathbf{u}}_{j|i}.\mathbf{v}_j$
       return $\mathbf{v}_j$
---

Figure 6: Routing algorithm used in capsnet

## B  Execution

### B.1  Public Repository Details

All code pertaining to this research paper has been hosted on Github at author's page: (https://raeidsaqur.github.io/CapsGAN/). All frameworks used and code execution information is available in README.md.

### B.2  GPUs Used

For execution, we used one multi-GPU rig (with 2 NVIDIA Titan Xps) and Google Cloud Computing instance with P1000 GPU.